\documentclass[10pt,twocolumn]{article}

\usepackage{geometry}
\usepackage{amsmath,graphicx}
\usepackage{dblfloatfix}
\usepackage{listings}
\usepackage[table]{xcolor}

\title{EASY REAL-TIME COLLISION DETECTION}
\author{J. Fabrizio\\LRDE-EPITA, 14-16 rue Voltaire, Le Kremlin-Bic\`{e}tre, France\\jonathan@lrde.epita.fr}
\date{}

\begin{document}

\maketitle

\graphicspath{{images/}}
\begin{abstract}
  This article presents an easy distance field-based collision
  detection scheme to detect collisions of an object with its
  environment. Through the clever use of back-face culling and
  z-buffering, the solution is precise and very easy to implement. Since
  the complete scheme relies on the graphics pipeline, the collision
  detection is performed by the {\it GPU}. It is easy to use and only
  requires the meshes of the object and the scene; it does not rely on
  special representations. It can natively handle collision with primitives
  emitted directly on the pipeline. Our scheme is efficient and
  we expose many possible variants (especially an adaptation to certain
  particle systems). The main limitation of our scheme is that it imposes
  some restrictions on the shape of the considered objects - but not on
  their environment.  We evaluate our scheme by first, comparing
  with the {\it FCL}, second, testing a more complete scene (involving
  geometry shader, tessellation and compute shader) and
  last, illustrating with a particle system.
\end{abstract}
\section*{keywords}
Collision detection, graphics pipeline, distance field

\section{Introduction}
\label{sec:intro}

Collision detection is a major problem in many domains: gaming,
simulation or augmented reality. We have to be able to detect when two
objects collide each other to allow interaction between objects and
realism in simulations and games. There are multiple formulations of
this problem. It can be the collisions of an object with the
environment or the collision among multiple sets of objects. Object
definition (implicit surface, mesh, composition of objects, etc.) and
properties of the object (rigid, articulated, deformable, etc.) lead
to various kinds of solutions. Due to user interaction and scenes
complexity, solutions must be efficient. The hardware used may vary:
{\it GPU} or {\it CPU} with or without parallelization.\\ Many methods
already exist and reviews are exposed
in~\cite{jimenez01,figueiredo02,kockara07,weller12}. We focus on
collision detection between an object and the environment. Many works
investigate other contexts (like~\cite{charlton19} that studies
collisions between multiple sets of objects) but it is out of the
scope of this article. Image-based collision detection has already
been investigated~\cite{wang12}. To detect collisions of one object
with the environment, we have to test collision of this object with
all the others. To be efficient, we have to reduce the number of tests
and simplify them.  To reduce the number of tests, the usage of
spatial partitioning representations have been used. The space is
partitioned and collision of objects are tested only if they belong to
the same part of the space. Many partitioning strategies have been
tested~\cite{held95}, octree~\cite{bandi95}, BSP tree~\cite{naylor92},
kd-tree~\cite{bentley79}. In case of animations, it is difficult to
keep the partition up to date (especially for BSP
tree~\cite{kumar99}). Equally, the usage of bounding volume
hierarchies have also been studied. The collision detection is sped up
by quickly discarding impossible collisions by testing simpler volumes
that encompass the objects. If the test fails with coarse bounding
volumes, collision is not possible. If the bounding volumes intersect,
objects may collide or may not. Finer bounding volumes are tested or
the objects themselves. This kind of methods need to process the
scene before. Many types of bounding volumes have been investigated:
axis-aligned bounding box (AABB)~\cite{bergen97},
spheres~\cite{hubbard96}, oriented bounding boxes
(OBB)~\cite{gottschalk96} and discrete-orientation polytopes
(DOP)~\cite{klosowski98}. To simplify collision tests, topological
methods have also been developed. A common strategy is to deduce
relative position of objects by projecting them on multiple different
axis like the Sweep-and-Prune algorithm does~\cite{cohen95}. Distance
field is another way to solve this problem. The idea is to compute a
distance field around the object and to compare these distance
fields. A solution similar to our scheme but restricted to the context
of cloth simulation is presented in~\cite{vassilev01}. The specific
case of clothes has received a special
attention~\cite{buffet19}. Multiple libraries are available to detect
collision (Bullet~\cite{bullet}, ODE~\cite{ode},
SOFA~\cite{sofa}, FCL~\cite{fcl, pan12}).\\
We offer in this article an easy
scheme to detect collisions between one object and the
environment. Our scheme relies on the rendering pipeline to perform
the detection: it is based on distance fields computed by the
rendering pipeline; this detection is then performed by the {\it GPU}
without additional development cost. Thanks to a clever usage of
z-buffer and backface culling the proposed method:
\begin{itemize}
\item is simple (its main advantage) but precise enough,
\item is easy to implement (relies on the rendering process and requires few lines of codes),
\item neither requires any knowledge on the geometry nor special
  representation or even preprocessing of the shapes: the meshes of the
  object and the environment is sufficient (and then is not memory consuming),
\item is able to manage primitives generated on the pipeline (by the geometry shader
  of the tessellation shader) or modified (compute shader or transform feedback...),
  the primitives remain into the VRAM,
\item is flexible and can manage reasonably some articulated/deformable objects
  and impose no restriction on the environment without additional cost.
\end{itemize}
The method can not manage however an object with (too many)
concavities (its main restriction).\\
In section~\ref{sec:principle}, we explain our collision detection
algorithm and we show multiple variants. In
section~\ref{sec:evaluation} we show some results. In
section~\ref{sec:conclusion} we conclude.

\section{Collision detection principle}
\label{sec:principle}

The goal is to estimate the location of the surface of the
object $S_o$ and the location of the surface of the environment $S_e$ to
be able to check if $S_e$ intersects the object or not. To estimate
the location of these two surfaces, we approximate them by two height
maps. To compute these two height maps, we project the object and the
environment onto the same plane. During the projection, we compute the
two height maps (expressed on the same plane). The comparison of these height
maps allows us to check if $S_e$ intersects the object or not. As they are both
expressed in the same plane, the comparison is trivial.

The graphics pipeline provides (by design) a simple, fast and
efficient way to compute these projections. We have to found a way to
compute simultaneously the height maps.\\

\begin{figure}[t]
  \centering
  a\hspace{-0.29cm}\includegraphics[trim=89 89 89 89,clip,width=0.20\linewidth]{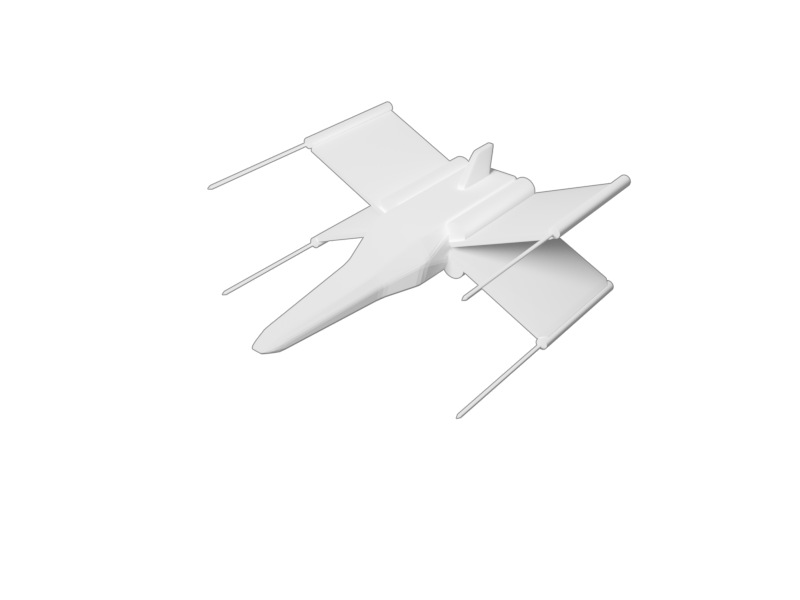}
  b\hspace{-0.29cm}\includegraphics[trim=89 89 89 89,clip,width=0.20\linewidth]{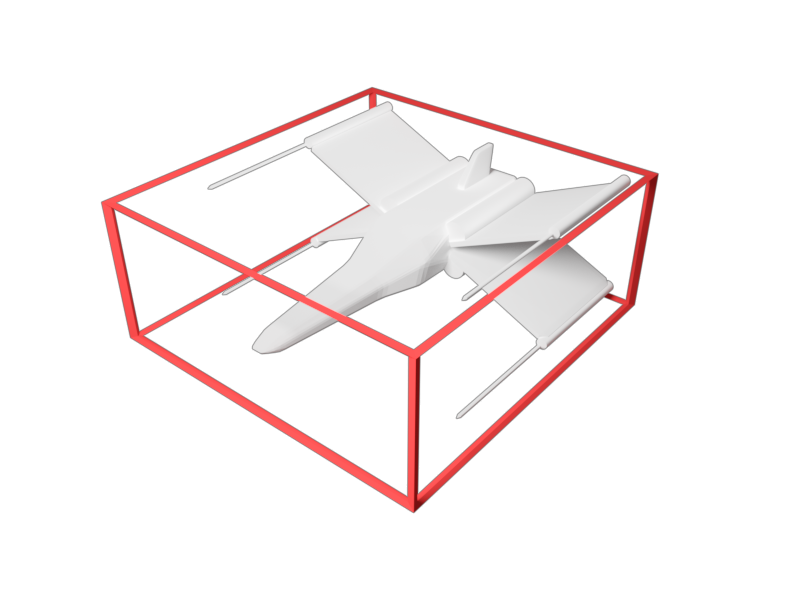}
  c\hspace{-0.29cm}\includegraphics[trim=89 89 89 89,clip,width=0.20\linewidth]{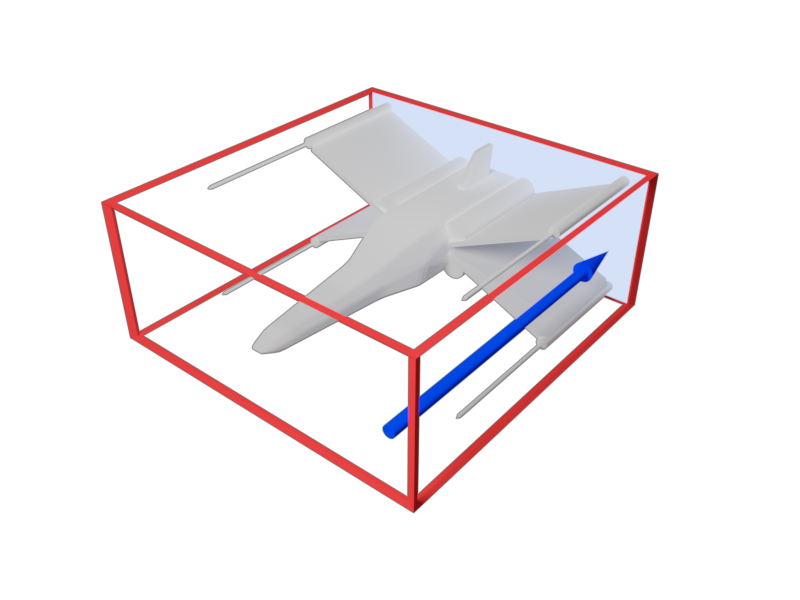}
  d\hspace{-0.29cm}\includegraphics[trim=89 89 89 89,width=0.20\linewidth]{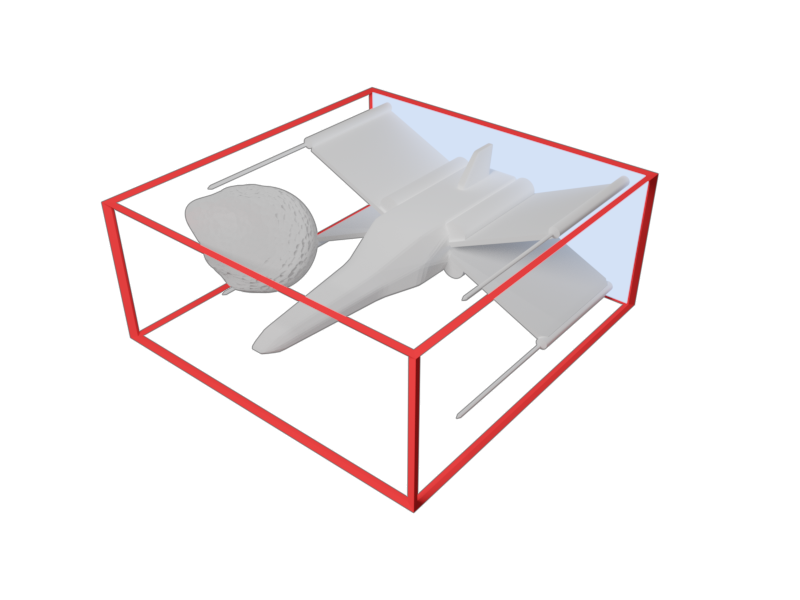}
  e\hspace{-0.29cm}\includegraphics[trim=89 89 89 89,width=0.20\linewidth]{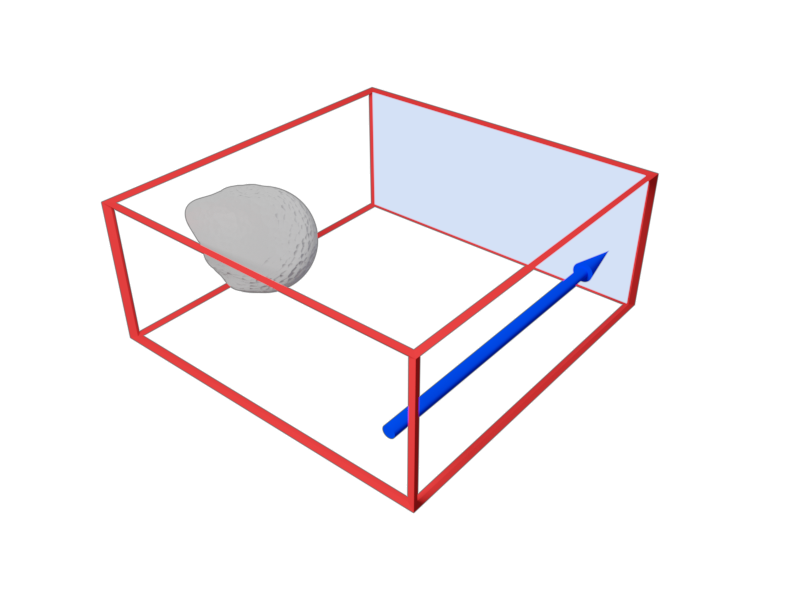}\\
  \vspace{0.1cm}
  f~\includegraphics[width=0.28\linewidth]{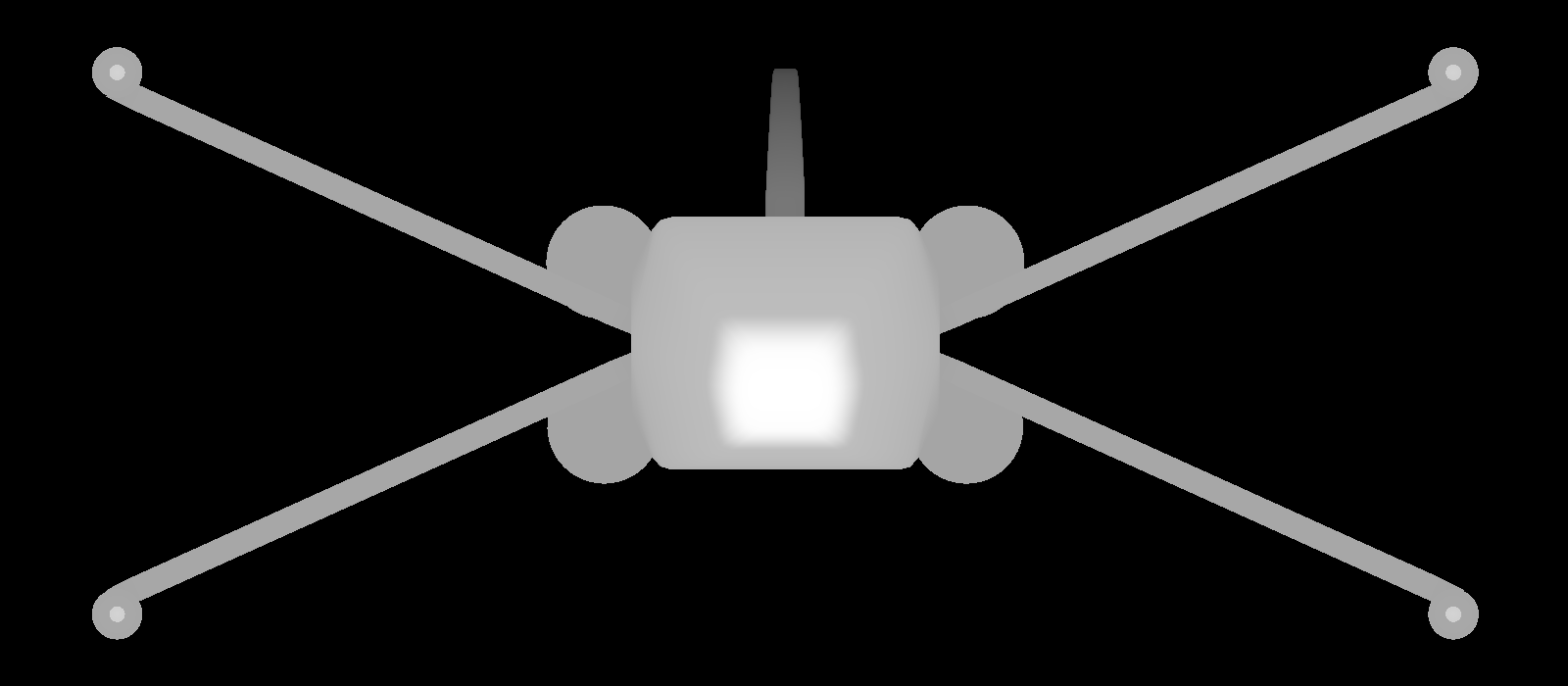}
  g~\includegraphics[width=0.28\linewidth]{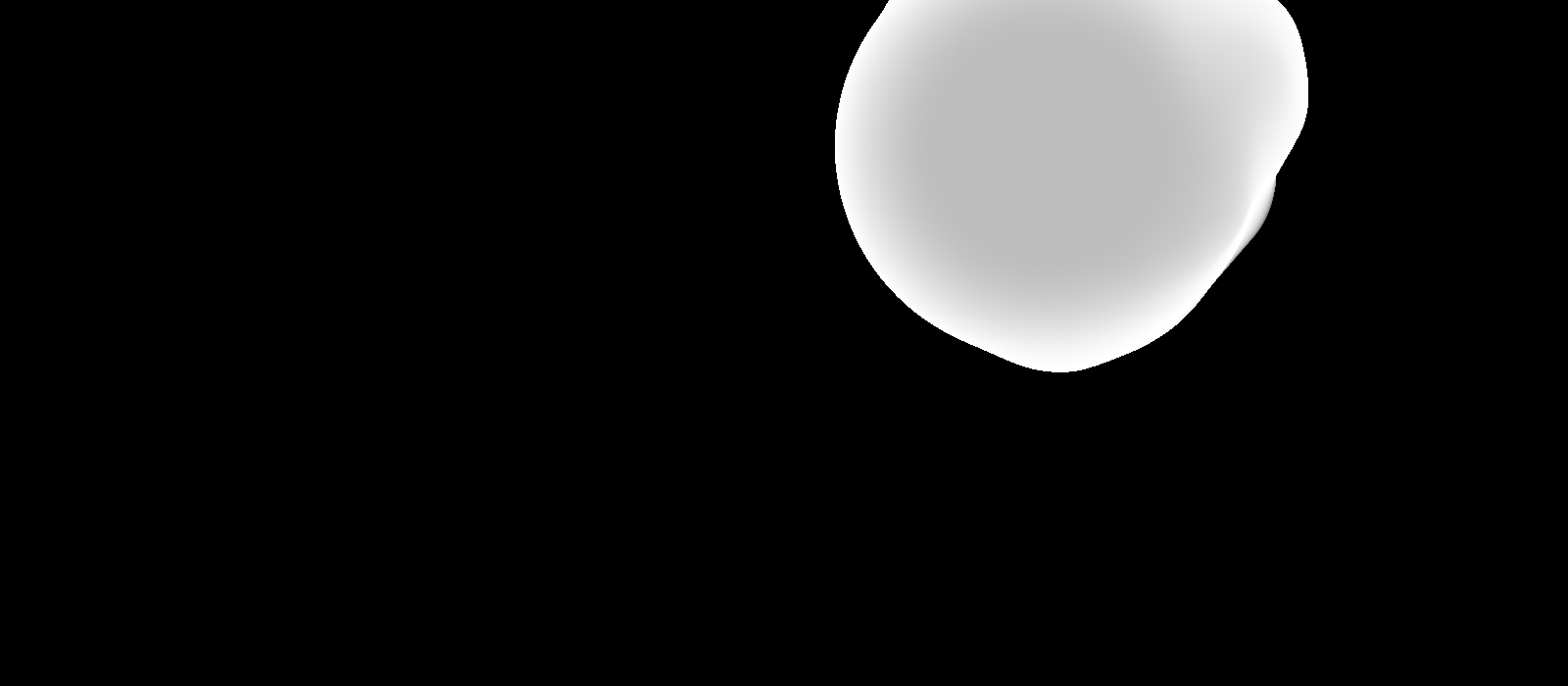}
  h~\includegraphics[width=0.28\linewidth]{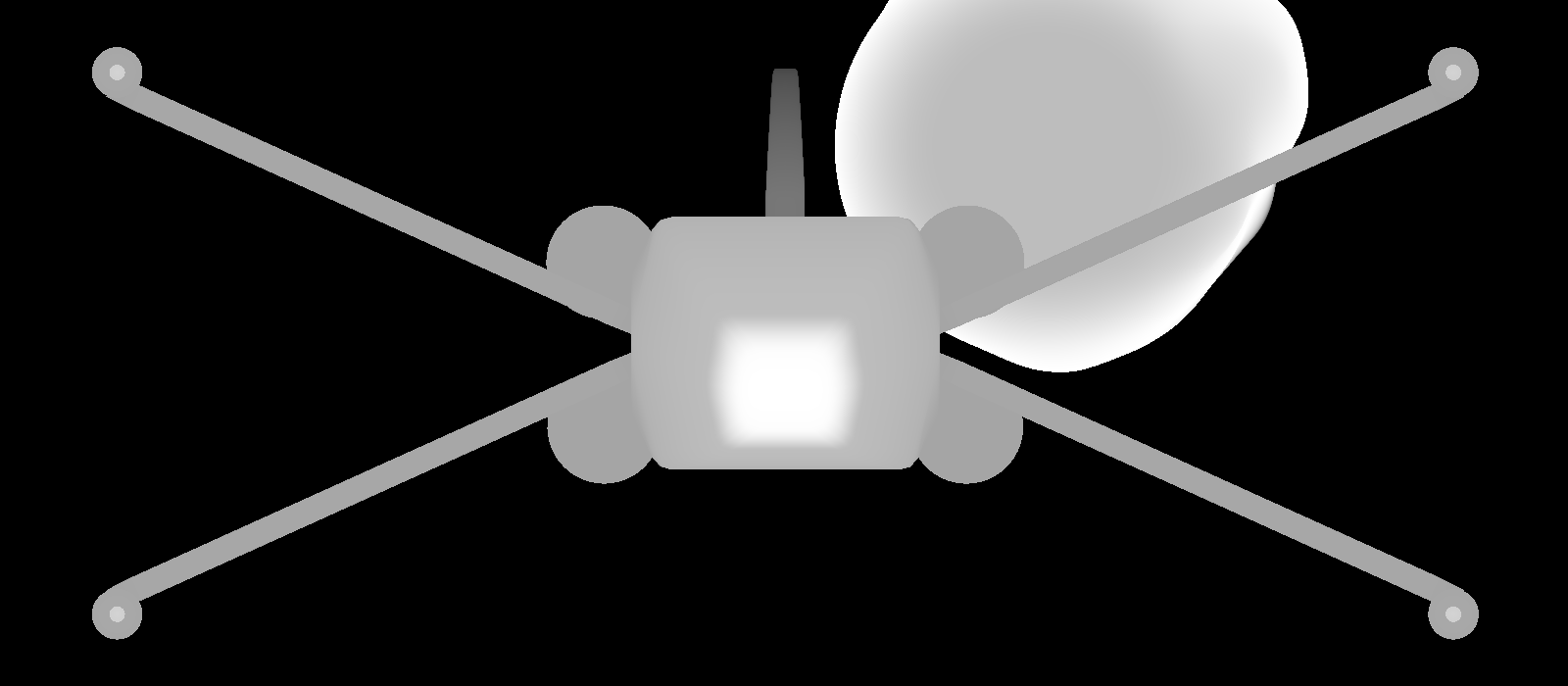}
  \caption{\label{fig:collision_detection_principle}\small a. The object, b. the
    bounding box $RC$, c. the projection of the object, d. the object in its
    environment, e. the projection of the environment onto the rear
    plane. f. the object's height map $S_o$, g. the environment's height
    map $S_e$, h. superposition $S_o$ and $S_e$.}
\end{figure}
{\bf The initial algorithm:} considering the object
Fig.~\ref{fig:collision_detection_principle}.a. We want to detect
collisions of the object (our x-wing) with the environment. Collision
detection is restricted to a bounding volume that encompass the
object; the red wire rectangular cuboid $RC$
(Fig.~\ref{fig:collision_detection_principle}.b).  We want to compute
the height map that approximate $S_o$. Then we project (only) the
spaceship onto the rear plane (according to the blue arrow -
Fig.~\ref{fig:collision_detection_principle}.c). To do so, we render
this object using an orthographic projection limited to $RC$. The
projection, the 3D clipping, etc. are natively performed by the {\it
  GPU}. This projection does not provide the height map directly but
during the rendering, the height map is automatically computed and the
result is in the content of the depth buffer after the
projection. However this rendering must produce a height map that
approximate the exterior surface of the object to watch. We want the
front part of the spaceship projected onto the rear plane and what we
get is the back of the spaceship.

We then perform the rendering in a particular way. As we want the
exterior surface, further from the rear plane: 1/we invert the z-buffer
(to get the furthest triangles and not the nearest), 2/we invert the
backface culling to compute the backface triangles.  The output (the
content of the depth buffer) is now the height map
(Fig.~\ref{fig:collision_detection_principle}.f) that describes $S_o$
(the front of the spaceship). These two tricks are the keystone of
the algorithm the get easily this height map.

The useful
part to deduce the height map is the content of the depth buffer. The
result of the rendering is useless. There is
then no texture mapping, no illumination computation, no special
effect, etc. Furthermore, as the projection is restricted to one
object and limited to a tiny part of the space that encompass this
object ($RC$): the rendering is then simple and very fast.\\

Now, to know if our
object collides with the environment, we have to study the space in
$RC$ (Fig.~\ref{fig:collision_detection_principle}.d). To do so, we
perform the rendering of the environment (without the considered
object) restricted to $RC$, using the same image plane and the same
orthographic projection
(Fig.~\ref{fig:collision_detection_principle}.e). As the environment
is facing the spaceship, the z-buffer and the backface culling are
used in the regular way contrary to the previous rendering. Again
textures, illumination, special effect, etc. are useless and the
rendering is limited to $RC$, it is very fast. The projection is the
same: no modification are needed between the two renderings. As
before, only the depth is required. The content of the depth buffer provides
the height map of the environment. We get a second height map, $S_e$,
the height map of the environment
(Fig.~\ref{fig:collision_detection_principle}.g).\\

The last task is
to compare the two height maps to detect whether there is a collision
or not (Fig.~\ref{fig:collision_detection_principle}.h). Thanks to our
previous tricks, the two distance fields have the same size and are
exactly in the same coordinate space, this makes it easy to compare
them very precisely. The comparison of the two depth maps
consists in checking (for every {\it pixel}) that the value of the
depth at the considered {\it pixel} in $S_e$ is lower than the depth
of the same {\it pixel} in $S_o$. This comparison can be performed
using a compute shader to ensure that all the data remain on the {\it
  GPU} side. The fragment shader can also perform this comparison
if we provide the first height map as a texture to the used fragment shader
for the second rendering.\\

\begin{figure}[t]
  \centering
  a\hspace{-0.29cm}\includegraphics[trim=95 100 100 05,clip,width=0.30\linewidth]{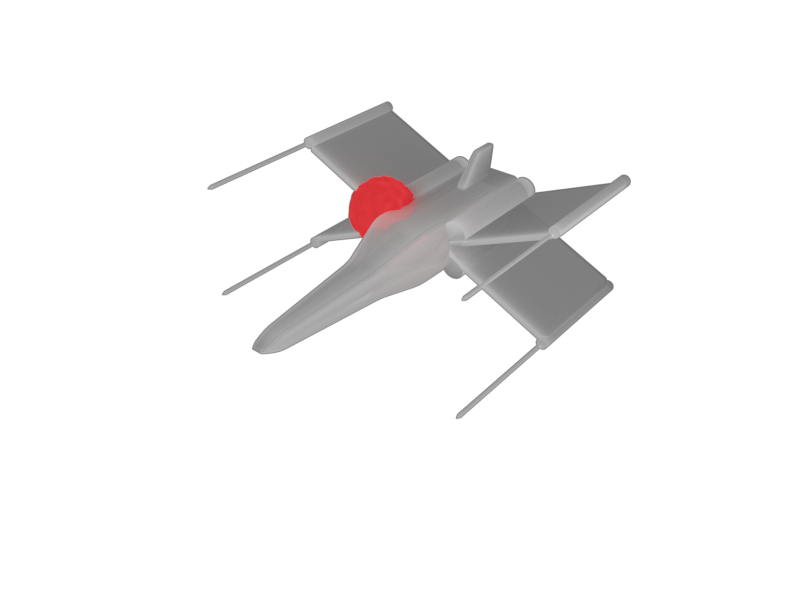}
  b\hspace{-0.29cm}\includegraphics[trim=95 100 100 05,clip,width=0.30\linewidth]{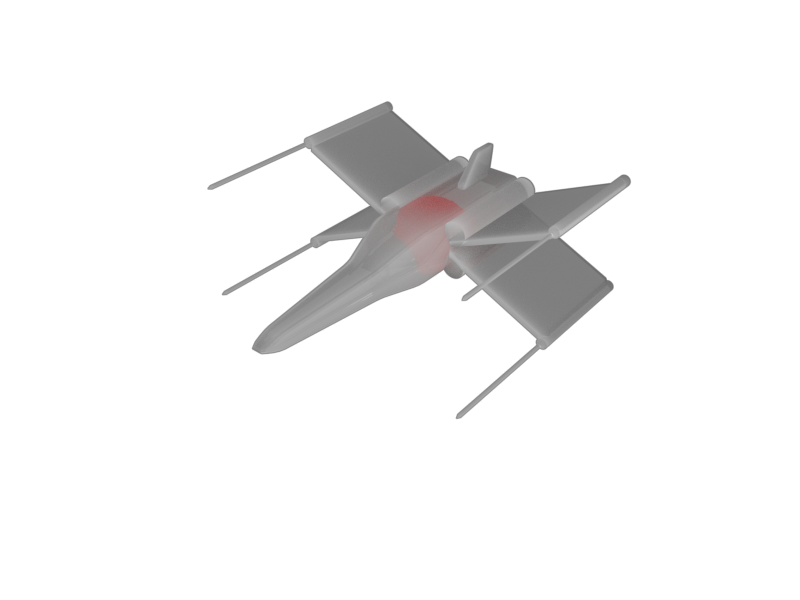}
  c\hspace{-0.29cm}\includegraphics[trim=95 100 100 05,clip,width=0.30\linewidth]{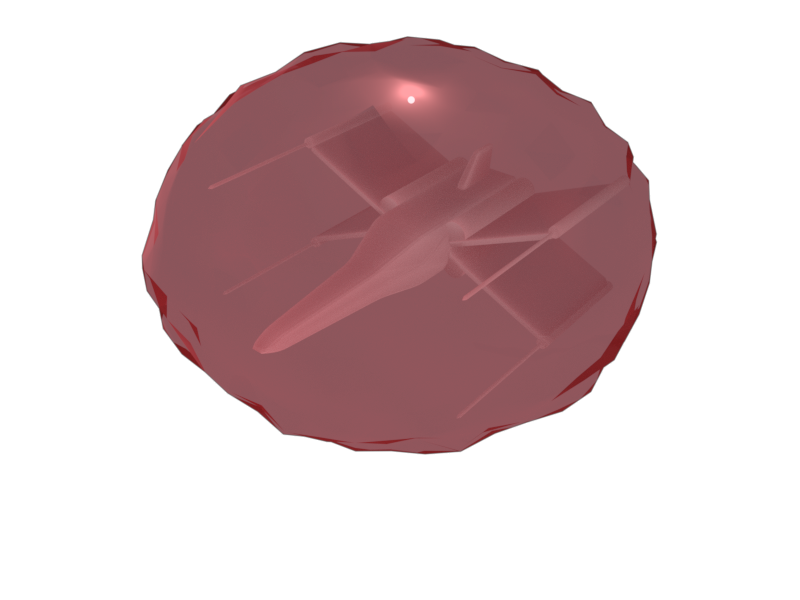}
  \caption{\label{fig:collision_capabilities}\small Our method detects (a) intersection, (b) inclusion
    of an object from the environment but not (c) when our object is encompassed by an object of
    the environment (c).}
\end{figure}

This algorithm is easy to set up and the major part of
the work is carried out by the rendering pipeline. This algorithm is
then simple to develop and fast to compute. It does not require any
knowledge on topology of the models but only the meshes. There are
three steps: 1/ rendering the object to get the height map of $S_o$
(with the depth buffer and the back face culling inverted). 2/ Every
time collision detection is needed, rendering the environment to get the
height map of $S_e$, 3/ compare the two height maps. If there is no
deformation of the object, $S_o$ is computed once and for all. If the
object changes, the render must be done to update $S_o$. The collision
detection is performed using just one simplified rendering.
This algorithm is able to detect when the two surfaces $S_e$ and $s_o$
intersect or when a part of the scene in totally included into the
object. On the contrary, this algorithm is not able to detect if the
object is totally included in an element of the environment (Fig.~\ref{fig:collision_capabilities}).\\

{\bf Strengths and weaknesses of the initial algorithm:} the major
strength of this algorithm is its simplicity as we have a collision
detection algorithm in very few lines of code but this algorithm is
also competitive in term of efficiency.\\

This algorithm is simple because the major of the work is performed by
the graphics pipeline. It does not require any additional development
to be run on a {\it GPU} - at least to compute the two height maps. It
is simple also because it only requires the meshes of the object and
the environment to generate the two height maps (and no other
model/representation) - no more than what we need to generate the
image. It is a challenge to solve a difficult problem in so few lines
of code.\\

It is reasonably fast because the graphics pipeline is well optimized,
by design, to perform these projections. Furthermore, contrary to classical
rendering, neither texture nor illumination need to be computed. It is efficient
because usually, in collision detection, we try to avoid unnecessary
intersection computation to save time. In the pipeline, it is
implicitly and natively carried out by many optimizations: clipping,
backface culling, etc.\\

Another advantage is that we can manage primitives emitted directly
onto the pipeline (by the geometry shader or by the tessellation stage).
It can also manage modifications of the primitives performed by the
graphics card (by the compute shader or by a transform feedback). This
is a huge advantage because we can compute the collision detection
without data transfer back to the RAM.\\

One can argue that the usage of the depth buffer is not precise enough
but: Firstly we use a linear depth (according to the orthographic
projection).  Second the maximal depth of the projection is the length
of $RC$. This length is commonly very low, the two computed distance
fields have a high precision: depth buffer's precision depends on the
distance between the near and far planes. As we limit the projection
to the length of the object, this distance is very low (supposed much
lower than the size of the scene) then we are very precise.\\

The major problem with this algorithm is that it is not applicable
with an object that has too many concavities. You can easily manage
convex objects or at least you can manage objects that have concavities
in the direction of the projection, but you can not manage objects with
arbitrary concavities.  The chosen spaceship is a {\it collision
  detection friendly object} as only one projection is required to
detect collisions and there is hardly no concavity in direction other
than the direction of the projection. Not all objects are so
simple.

Note that even if they are limitations on the shape of the
object, there is no restriction on the shape of the environment. It
can have concavities. There is also no restriction also on the evolution
of the environment, you can manage moving part or deformable environment
without additional cost or complexity because $S_e$ is recomputed at
each iteration.\\

\begin{figure}[t]
  \centering
    a\hspace{-0.29cm}\includegraphics[trim=5cm 5cm 5cm 5cm,clip=true,width=0.30\linewidth]{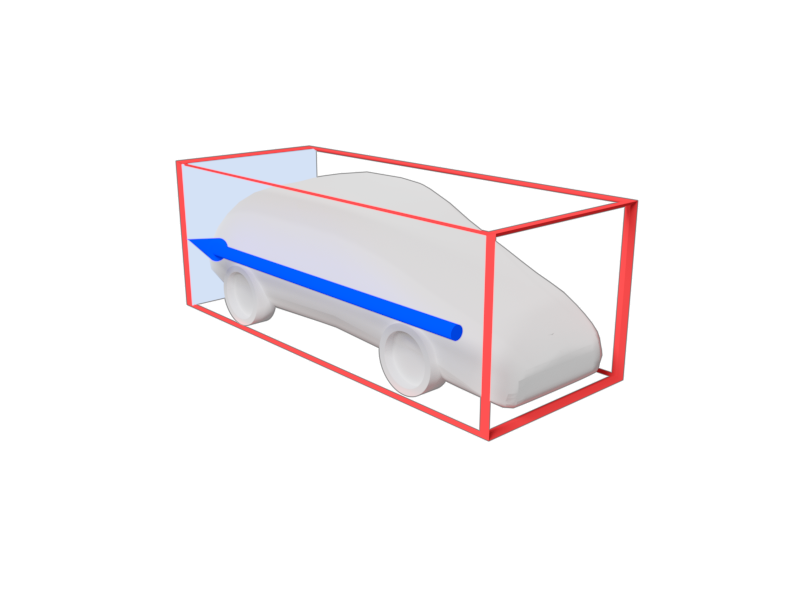}
    b\hspace{-0.29cm}\includegraphics[trim=5cm 5cm 5cm 5cm,clip=true,width=0.30\linewidth]{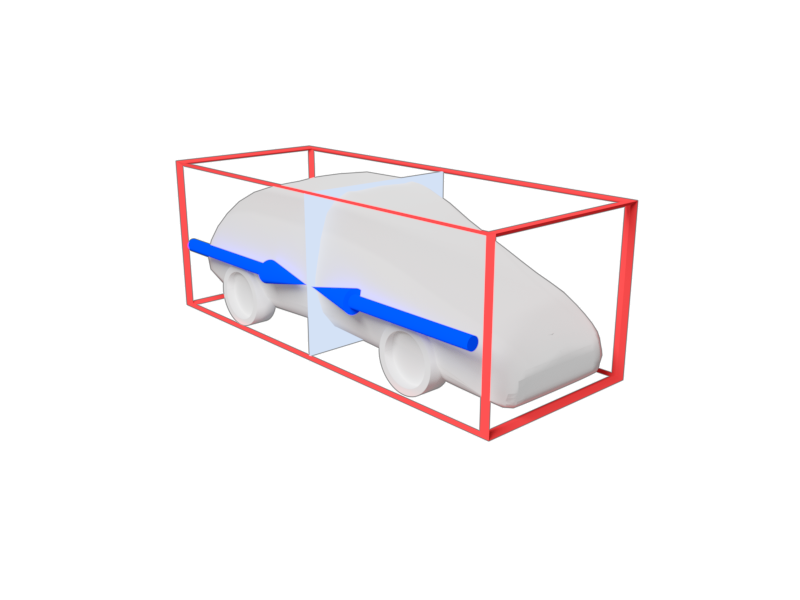}
    c\hspace{-0.29cm}\includegraphics[trim=5cm 5cm 5cm 5cm,clip=true,width=0.30\linewidth]{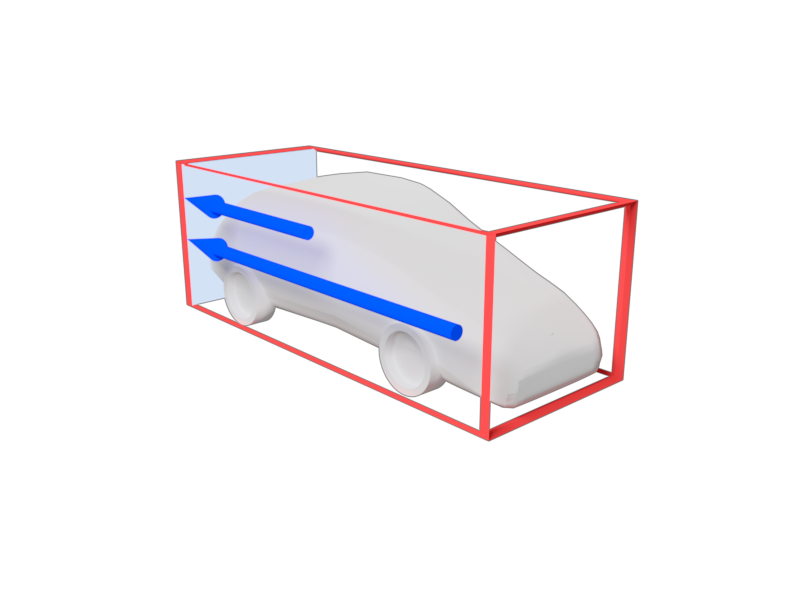}\\
    d\hspace{-0.29cm}\includegraphics[trim=5cm 5cm 5cm 5cm,clip=true,width=0.30\linewidth]{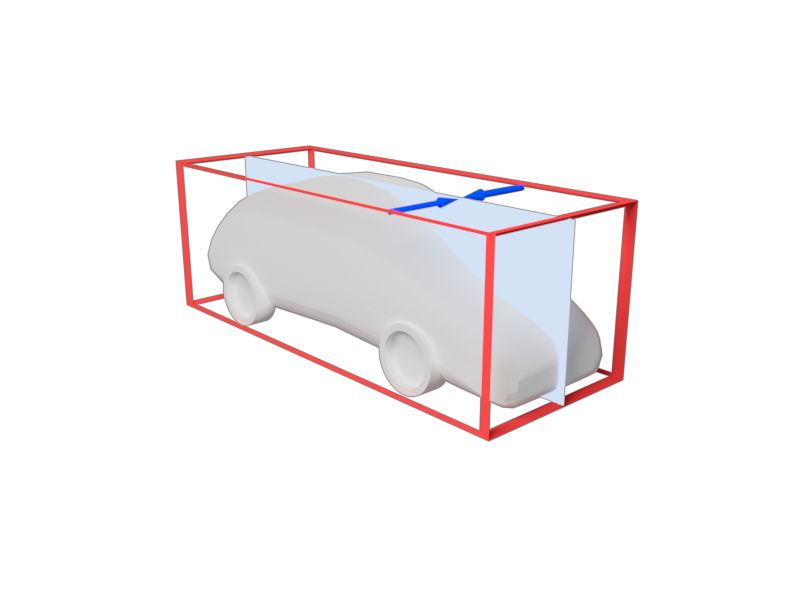}
    e\hspace{-0.29cm}\includegraphics[trim=5cm 5cm 5cm 5cm,clip=true,width=0.30\linewidth]{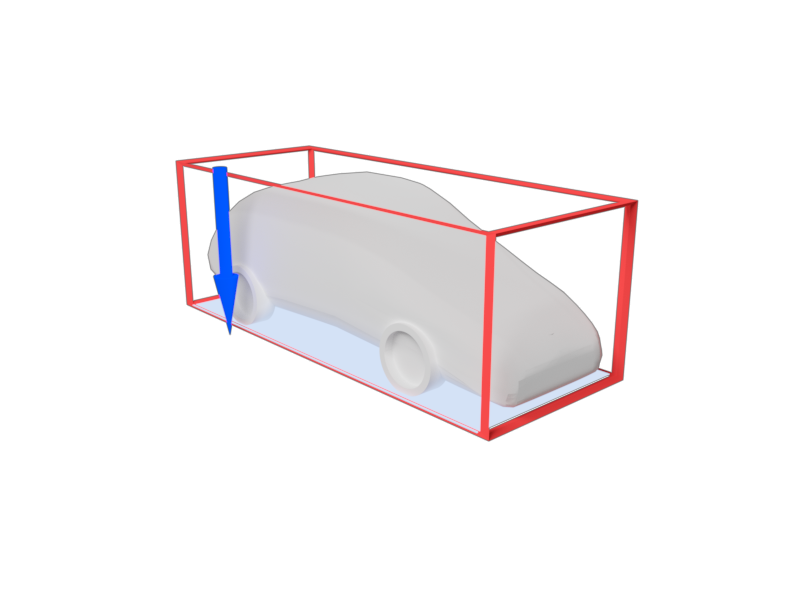}
  \caption{\label{fig:collision_detection_car}\small Various projections
    for an object.}
\end{figure}

There is a preliminary work to decide which direction(s) to use for
the projection(s). It is important to select the correct one(s) to
minimize the number of renderings.  If you consider the car
Fig.~\ref{fig:collision_detection_car}. The projection onto the rear
plan (a) as for the spaceship is not adapted as rear collision won't
be correctly managed. An idea is to compute two height maps instead of
one (b). The first one to detect collisions with the front part of the
car, the second one to detect collisions with the rear part of the
car. In this case, at runtime, this obliges to project the environment
twice. In this specific case were the two projections have the same
axis, it is possible to improve the process by projecting twice the
object but on the same plan (c). One to get the height map of the
surface of the front of the object, the second to get the height map
of the surface of the rear of the object - this provides two
boundaries. The difference between the two renders is that we have to
invert the backface culling and the z-buffer between them. With these
two height maps, we can detect a collision by rendering the
environment once and checking if $S_e$ is encompassed between the two
boundaries. In practice this solution imposes some constraints onto the
environment, and it is then not easy to use. Another, solution (d) is
equivalent to (b) but the size of the image plane is bigger (b). The
last solution (e), which is better, is to project the
car on the floor. This allows to detect collisions by performing only
one projection in the direction of the floor. For all these solutions
we still have a problem to detect properly collisions with the car
wheels.

This example shows that incorrect
direction leads to incorrect collision detection and/or to increase
uselessly the number of renderings.\\

{\bf Algorithm variations:} the main advantage of this scheme is its
simplicity. You have, under certain circumstances, a simple solution -
in few lines of code - to a complex problem. Another advantage is
its versatility. We briefly list some of possible
variants/improvements.\\

{\it Stencil buffer:} we may
save time by using a stencil buffer. On
Fig.~\ref{fig:collision_detection_principle}.f, the area covered by the
spaceship is very low compared to the area of the complete image. To
compute $S_e$, we do not need to evaluate positions not covered by the
spaceship. We can deduce from $S_o$ a stencil to restrict the
computation of $S_e$. In practice, the improvement brought by the
usage of the stencil buffer may vary depending on the implementation
as frequently, the stencil buffer is applied late in the pipeline.\\

\begin{figure}[t]
  \centering
  \begin{minipage}{0.61\linewidth}
      \centering
  \hspace{-2cm}\includegraphics[trim=89 89 89 89,clip,width=0.49\linewidth]{my_xwings_02}\\
  \vspace{-0.8cm}~~~~~~~~~~~~~~~\includegraphics[trim=89 89 89 89,clip,width=0.49\linewidth]{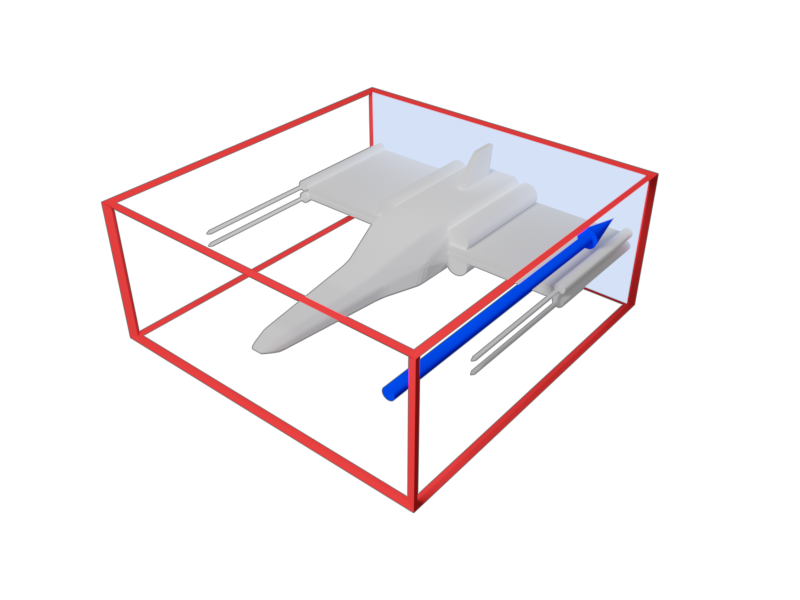}\vspace{-0.3cm}
  \caption{\label{fig:Space_ship_wings_moving}\small Spaceship with moving parts.}
   \end{minipage}
  \begin{minipage}{0.325\linewidth}
      \centering
  \hspace{-0.3cm}\includegraphics[height=2.0cm]{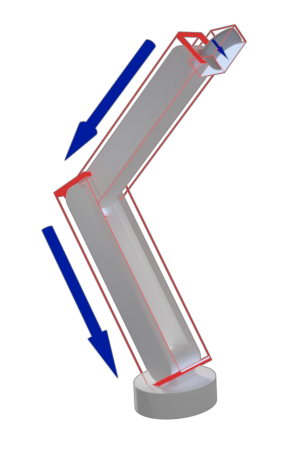}\\
  \includegraphics[height=2.4cm]{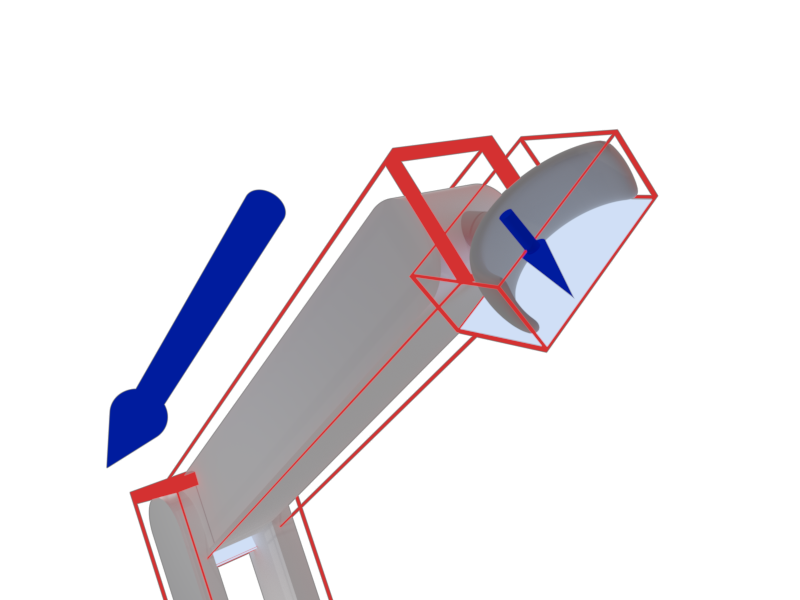}
  \caption{\label{fig:collision_detection_bot_01}\small Projection
    computation for an articulated object.}
   \end{minipage}
\end{figure}

{\it Deformable object:} to manage an object that
changes its shape over the time, we have to update $S_o$ at every
iteration. In the case of the spaceship if, like an x-wings, the wings
move (Fig. \ref{fig:Space_ship_wings_moving}), $S_o$ must be
recomputed. The limit of this strategy depends on the shape of the
object and the movement. We can easily recompute $S_o$ as long as we
can determine the direction of the projection and the deformation does
not bring concavities. Once again, if the object is compliant with our
scheme, modification on the shape of the object is simple to
handle.\\

{\it Articulated object: } most of the time, we can not handle
all the parts with only one projection. A solution is to manage each
part independently as in Fig.~\ref{fig:collision_detection_bot_01}: a
projection is computed for each part of the arm and for the hand.\\

{\it Particle system:} if the particles are in a limited part of the
space and if this zone is simple enough (convex, star convex), it is possible to use our
scheme to manage collisions between the particles and the environment
(not between particles). In this case, $S_e$ is computed in the
direction of the propagation of the particles and all the particles share
the same height map for $S_e$. In case of multiple directions, it is possible to
compute $S_e$ in all 6 directions of a cube centered at the centroïd
of all particles. As the particles share the same height map, $S_e$ is
computed only once per direction regardless of the number of
particles.

Note that in every direction, you are limited to the
first visible obstacle. You can manage more obstacles with
depth peeling for example, but it is
much more time consuming.\\

{\it Interaction:} when a collision is detected, it is useful to know
additional information on the collision. With this scheme it is for
example easy to detect which object of the environment collides with
$S_o$. When computing $S_e$, it is possible to simultaneously store the
{\it ID} of every object of the environment in a separate
buffer. During the rendering, we keep not only the depth, but also the
{\it ID} in a render buffer. When an object collides, we know, by
querying this buffer, the {\it ID} of the object. Another useful
information to compute interaction is the normal of the surface at the
contact point. As with the {\it ID} recording, it is also possible to
record the normal of $S_e$.  This is very useful to compute
interactions like bounce of particles (even if, it is difficult to
integrate all the normals in every contact points; when the object
collides with the environment, there is not a unique point of contact
but a set of points with many different normals. Moreover, it can have
multiple sets of contact points). We can store other information
(like force fields) to manage interactions with the environment.
Used with particles, it can be very useful.

\section{Evaluation}
\label{sec:evaluation}
\begin{table*}[b]
  \begin{center}
  \scriptsize
  \begin{tabular}{|c||c|c|c|}
    \hline
    & The {\it FCL} & Our method & Student's t-test \\
    \hline\hline
    Average time on all frames (s) & 0,0002050846275 & {\bf 0,00010161438625} & p-value=2.929e-08 (significant at $1\%$)\\
    \hline
    Average time on frames with collision (s) & 0,000139403807767 & 0,000103404759223 & p-value= 0.09391 (non-significant)\\
    \hline
    Average time on frames without collision (s) & 0,000323771021053 & {\bf 9,8379150877193E-05} & p-value= 5.535e-11 (significant at $1\%$)\\
    \hline
    Precision on detection & {\bf +} & - & \cellcolor{gray!25}\\
    \hline
    Memory usage (coarse estimation) & between 5 and 6 Mo & {\bf inf to 1 Mo} & \cellcolor{gray!25}\\
    \hline
    Easy to use & - & {\bf +} & \cellcolor{gray!25} \\
    \hline
    Can manage primitives emitted/modified & - & {\bf +} & \cellcolor{gray!25} \\
    by the video card & & & \cellcolor{gray!25}\\
    \hline
  \end{tabular}
  \caption{\label{tab:evaluation}\small Comparison of the two
    methods (The FCL and our). Comparisons on time (with statistical hypothesis testing
    to check the representativeness of the difference of the two
    means), memory consumption, precision and usage simplicity.}
  \end{center}
\end{table*}

\subsection{Quantitative results}
{\noindent\textbf{Comparison with the FCL:}\quad} Our method is very easy
to use however we have to validate that the method is competitive. To
do so, we compare, for the same scene, the collision detection from
our method and the collision detection using {\it The Flexible
 Collision Library}~\cite{fcl,pan12} (We have chosen the FCL because
it is common, fast, easy to use, precise and popular enough). We use
a sequence of 800 frames: our spaceship is flying into a tunnel with
multiple obstacles. The spaceship contains $9\:716$ triangles and the
environment contains $69\:842$ triangles. Table.~\ref{tab:evaluation}
summarizes the result of the comparison.\\

The first important criterion is the speed. To be
precise in the measurement, we measure the time to draw the scene with
and without collision detection, and then we deduce the time spent to
compute collision detection. The measure is averaged on the 800
frames. The time spent by the {\it FCL} to initialize
data structures is not measured; we compare only the rendering
time. The CPU used (by the {\it FCL}) is a 4.00GHz i7-6700K CPU. The
GPU used (by our method) is a GeForce GTX 1650. The model of the
spaceship and the environment is not modified during the test (no
deformation/no moving part).
According to table~\ref{tab:evaluation} on average our method is
faster than the {\it FCL}. We check the representativeness of the
comparison with a Student's t-test and validate that
 the comparison is relevant. Moreover, we show that the executing
time is more stable for our method compared with the {\it FCL} which
vary depending on whether there is a collision or not.\\

Unfortunately,
we use only one projection to detect collision, this means that the
detection may generate a false positive if a small object is right
behind the spaceship (between the spaceship and the rear plan). In
practice, we counted 4 errors over the $800$ frames (at always the
same situation: an obstacle passing throw the X-wing and when goes out
at the rear of the X-wing, our algorithm detects with one frame late
that the obstacle is out). We can then consider that the {\it FCL}
is more precise (however we could have use two projections instead
of one).\\

Using system {\it mstat} we have a coarse
estimation of the data allocated by the {\it FCL}, and we measure more
than $5$Mo. It is much more compared to our two small height maps (less
than 1Mo) stored into the GPU card.\\

To finish the comparison, our method is much more
easy to use because it simply relies on the rendering process. For example,
we can easily update the detector if the spaceship moves its wing.
It is not so easy with the {\it FCL}.
Our method is then easy to use but also competitive.\\

{\noindent\textbf{Test in a more complex environment:}\quad} To measure the
usability of our method in a real scene we test our
method on a more complex and textured scene (Fig.~\ref{scene_completed}.a).
This environment is composed of more than $315\:000$ triangles.
The object is still a flying X-wing (Fig.~\ref{scene_completed}.b) but now
with $16\:792$ triangles. There are various situations represented in this scene.
An object of the scene is continuously moving (Fig.~\ref{scene_completed}.c).
To fully measure the benefit of our method, in addition to the 315K triangles of the environment,
some parts of the scene are generated by a geometry shader or by a tessellation shader: the tessellation
shaders generate and animate a flag (Fig.~\ref{scene_completed}.d) while the
geometry shader generates green plants (Fig.~\ref{scene_completed}.e).
These primitives remain into the graphic card. To finish, we
add a firework composed of $59\:000$ particles (Fig.~\ref{scene_completed}.f). These
particles are updated (color and position) by a compute shader between two draws. Again, these
primitives never go back to central memory and remain into the graphic card.

The primitives used for the scene are the triangles. However, the output of the geometry
shader is a collection of lines and the particles of the firework are drawn by
simple colored points. This demonstrates another advantage of the proposed solution:
we can detect intersections with other primitives (lines) and even test the inclusion
of another primitive in the object (if the particle hits the object).

We can not compare with the {\it FCL} as previously because we
never store the primitives in the RAM (this is a huge advantage in
favor to our method) and even we can test intersection/inclusion with different types of primitives
(triangles, lines, dots). As this comparison is not possible easily, we will measure the
impact of our method by measuring the average time to draw the frames of the scene, with and
without our collisions detection process.

Note: in practice, as we have multiple programs to draw the objects of
the scene, we link all programs again but with a simple and generic fragment shader
that simply draws the depth (we have a linear depth at this point).
The content of the fragment shader is very simple:
\begin{lstlisting}
#version 450

layout(location=0) out vec4 output_d;

void main()
{
  float depth = (gl_FragCoord.z+1.0)/2.0;
  output_d = vec4(depth, depth, depth, 1.0);
}
\end{lstlisting}
This is for the test but in practice, we could have simply stored
the ID of the object of the environment
in the render buffer (instead of the depth) to be able to know which object
intersects and read the depth directly from the depth buffer (this
works for the intersection with one object but may fail with two intersections
as the first object may hide the second one).

The scene is around $600$ frames long. We use an entry level video card
NVidia GT 1030 with $384$ graphic cores to test the
usability of our method. We measure the time to draw the scene. We average the time
to draw the frames of the scene: $0,0058$ second and the average time to draw and
to test the collisions: $0.0064$ second. As the two means are close
to each other, we have applied a Z-test to validate that the comparison
is relevant and the difference is statistically significant. Even so,
the difference is low. To improve the speed, one could use simplified
meshes or draw only some parts of the environment but these improvements are valid
whatever the method you use. Finally, this test shows that:

\begin{figure}
  \begin{center}
    a.\includegraphics[width=0.45\linewidth]{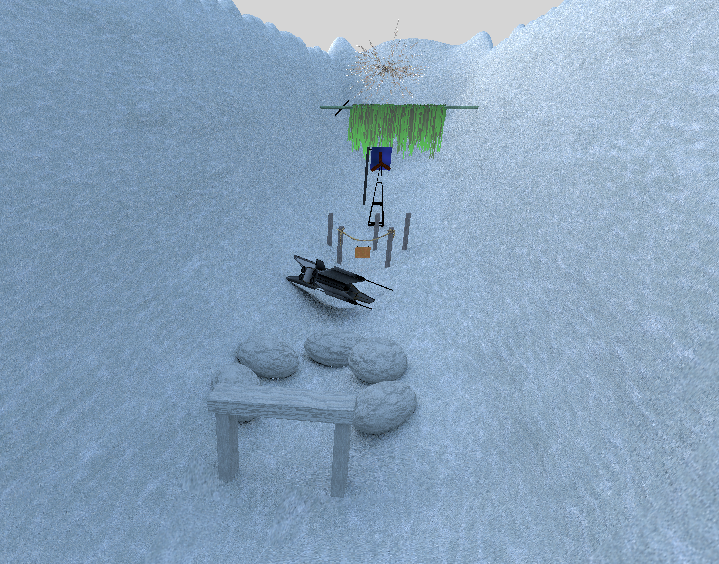}~
    b.\includegraphics[width=0.45\linewidth]{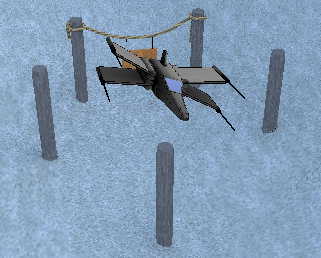}
    c.\includegraphics[width=0.25\linewidth]{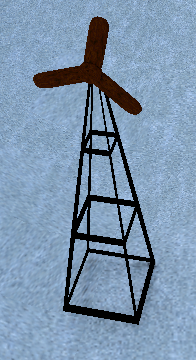}
    d.\includegraphics[width=0.25\linewidth]{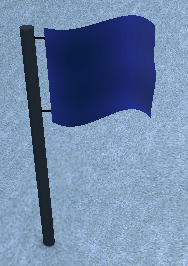}
    f.\includegraphics[width=0.25\linewidth]{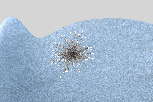}
    e.\includegraphics[width=0.75\linewidth]{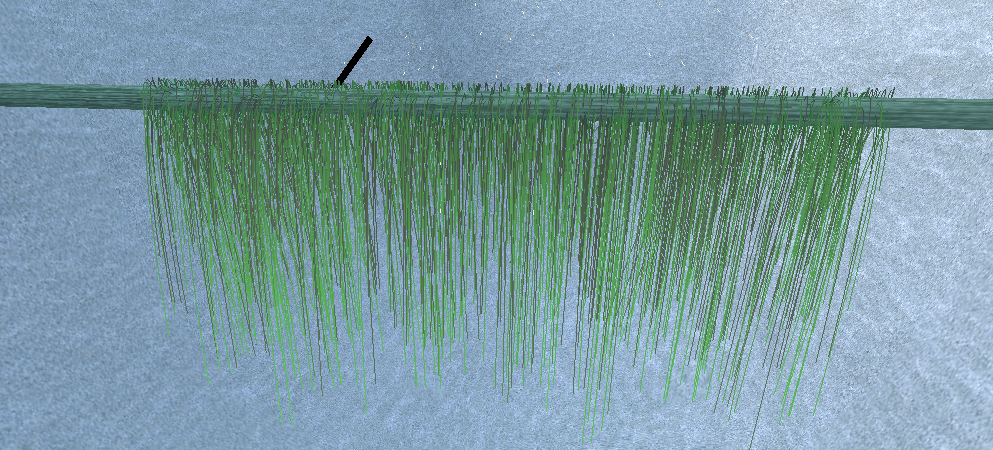}\\~~
    \caption{\label{scene_completed}\small a. The scene. b. The flying x-wings.
      c. A moving object. d. A flag, computed by the tessellation shaders.
      e. Plants, generated by the geometry shader (drawn with lines).
      f. Particles (dots) of the firework animated by a compute shader.}
  \end{center}
\end{figure}

\begin{itemize}
  \item we can manage moving and deformable objects in the environment,
  \item we can manage different types of primitives in the environment (points, lines and polygons),
  \item we can manage primitives emitted/modified directly on the pipeline,
  \item the time spent to compute the collision is very low compared to the time to draw the scene,
  \item the time to compute the collisions is reasonable.
\end{itemize}
This test is realized on an entry level card, this demonstrates
 even more the usability of the method. %

\subsection{Qualitative results}
\begin{figure}
  \begin{center}
    a.\includegraphics[width=0.85\linewidth]{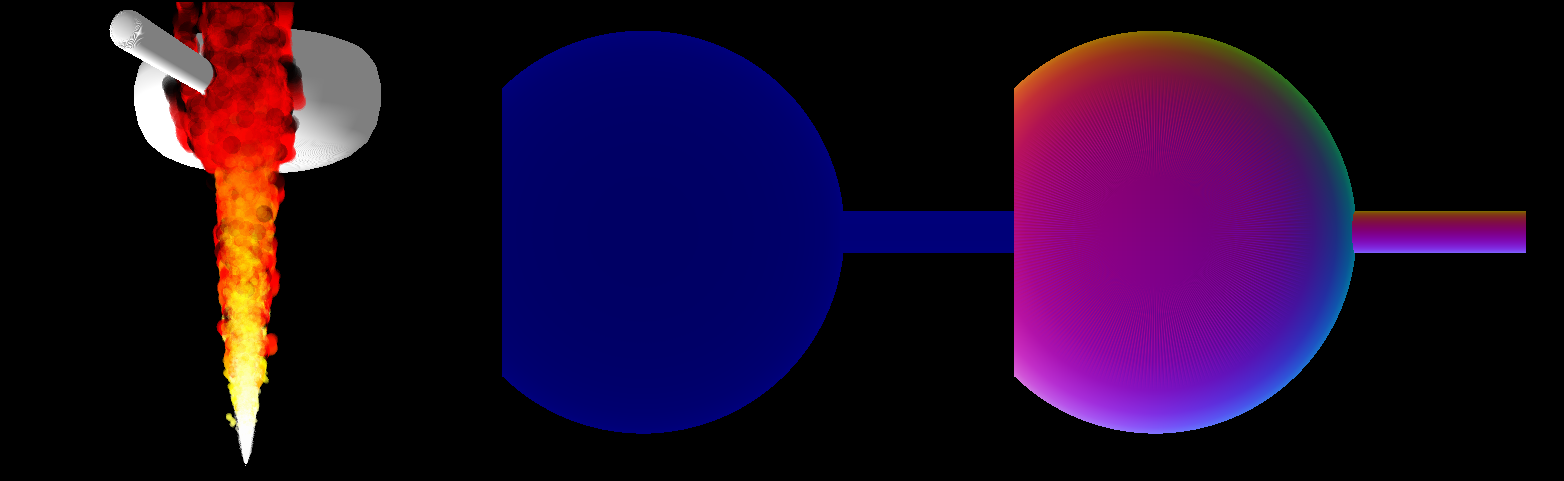}\\
    b.\includegraphics[trim=4cm 0cm 0cm 0cm,clip=true,width=0.25\linewidth]{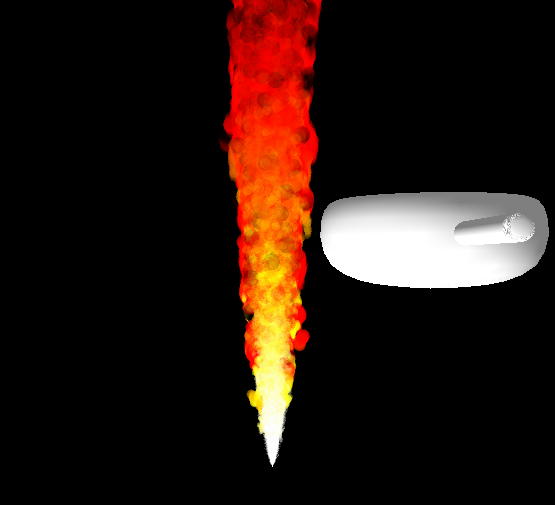}
    \includegraphics[trim=4cm 0cm 0cm 0cm,clip=true,width=0.25\linewidth]{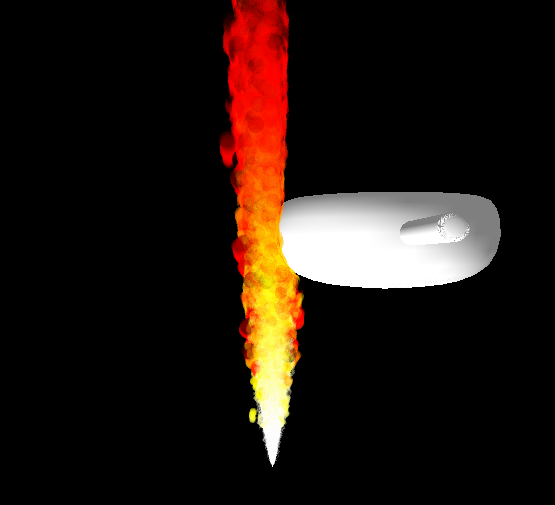}
    \includegraphics[trim=4cm 0cm 0cm 0cm,clip=true,width=0.25\linewidth]{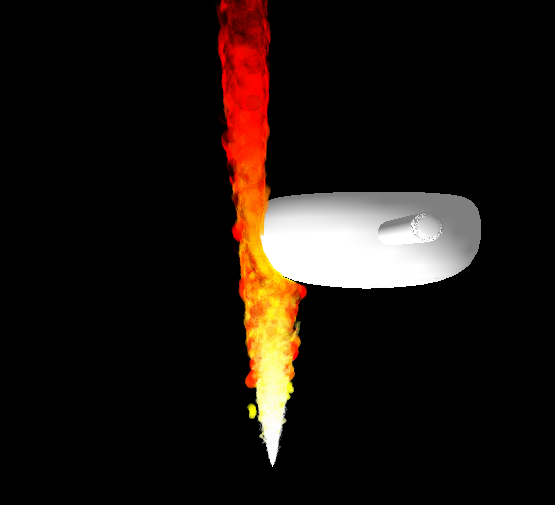}\\~~
    \includegraphics[trim=4cm 0cm 0cm 0cm,clip=true,width=0.25\linewidth]{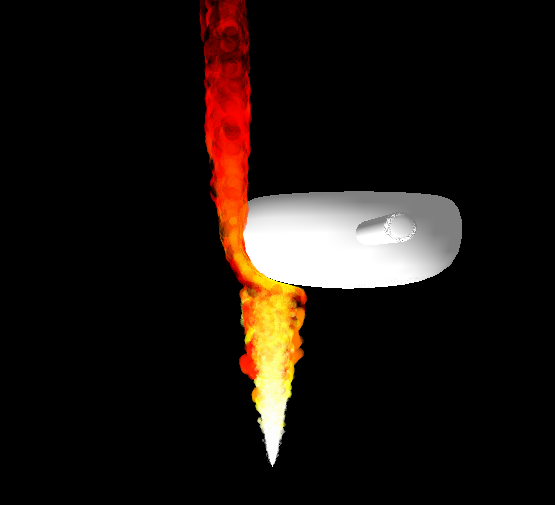}
    \includegraphics[trim=2.5cm 0cm 1.5cm 0cm,clip=true,width=0.25\linewidth]{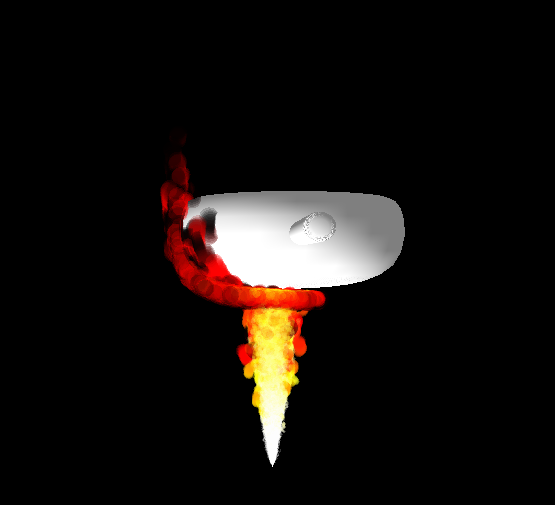}
    \includegraphics[trim=2cm 0cm 2cm 0cm,clip=true,width=0.25\linewidth]{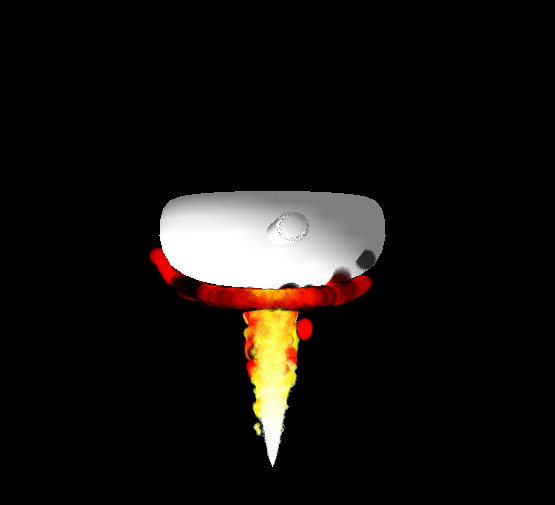}
    \caption{\label{particle_system_result}\small a. From left to right: The
      fire surrounding the skillet and the handle, the height map used
      to detect collisions, the normal of the object used to compute
      bounces.  b. The skillet is moving above the fire and the
      particles of the fire are deflected.}
  \end{center}
\end{figure}

We use a simple particle
system to simulate fire. Particles start from a point $S$
and go up with a perturbed motion and color variations. There are
around $50\:000$ particles represented by point sprites with varying
size. A skillet is modeled ($81\:920$ triangles) and the fire is
expected to be deflected by the skillet. To do so, the normals are
computed simultaneously with $S_e$. When a collision is detected, a
bounce is computed thanks to the normal at the contact point. $S_e$ is
computed from the particles' point of view, i.e. from the point $S$,
to the sky direction. A unique $S_e$ is computed for all the
particles at each iteration. The result is illustrated
Fig.~\ref{particle_system_result}.a: the fire is well deflected by the
object. In this figure the height map and the normals are
exposed. These two images are provided to the compute shader that
updates the particles to compute deflection. A sequence of motion is
shown Fig.~\ref{particle_system_result}.b. The collision detection in
mean takes around $2$ms with a $1024\times1024$ height map.  These
times prove the usability of the method. They vary depending on the complexity of the
scene and the dimension of the height map (a speed/precision trade
off). This test shows also the versatility of our method.

\section{Conclusion}
\label{sec:conclusion}

We have presented an easy collision detection scheme. This scheme is able
to manage collision detection between one object and its environment. The scheme can detect
if the surfaces collide or if an element of the environment is included in the object but
not if the object is included in an element of the environment. The main part of
the work is performed by the graphics pipeline. It relies on a clever
usage of the backface culling and the z-buffer (in an uncommon way). We
have exposed multiple variants/improvements to show that the scheme is
versatile, and we have illustrated it on a particle system.\\

The main advantage of this scheme is its simplicity. It does not require
anything else but the meshes. This means that any modification of the
meshes is managed without nearly any additional cost. We can handle easily
modifications in the model of the environment (moving parts, deformations)
as well as modifications in the object. Even if
our method is certainly not the fastest one, our method is efficient,
thanks to the usage of the graphics pipeline. A simple {\it
  POC} shows that our method is not only easy to use but consumes few
time on an entry level video card, is precise enough and very
efficient in terms of memory consumption. With the comparison against
the FCL, we validate that the solution is competitive.\\

Another strong advantage is that the algorithm can handle primitives
used for the environment, emitted directly by the pipeline (geometry
shader, tessellation process) or modified by the video card (compute
shader, transform feedback). This is a strong advantage because we
do not need to transfer all these primitives from the VRAM to the RAM
to test the collision with our object. There are a lot of elements
computed directly on the graphic card (the terrain by tessellation,
details by particles...), and we are able the handle natively all these
primitives.\\

The main limitation is that
it imposes strong restrictions on the shape of the object to
test. The object mustn't have concavities in directions different
from the projection direction (but we are not limited to convex
object). There are plenty of non-compliant objects, but we have
illustrated our process with a car, an X-Wing, a particle system: it
still remains many handled objects. With representative examples we have
shown that even with restrictions many objects can be handled.\\

There is no restriction on the environment for collision detection with
a fully compliant objects. The surface of the environment is projected no
matter its shape (even with concavities) and perfectly handled.\\

We rely on the depth buffer. The precision of the depth
buffer depends on the distance between the near and far
planes. As we limit the projection to the length of the considered object, this
distance is usually very low (much lower that the size of the scene) then
usually, we are very precise (we handle perfectly the asteroid Fig.1 in the
bounding box).\\

If applicable, the proposed algorithm is a strong solution - a simple
fast precise solution for compliant objects (or with slight imprecision
for near compliant objects). You have, in very few lines of code, a
competitive and simple solution to a complex problem. We have presented the
advantages but also the limitations to help the reader to know if applicable
in its situation.\\

We do not provide any algorithm for projections selection. We can choose
the correct direction manually during object modeling, but it would
be interesting, in the future, to investigate to find an automatic algorithm. This
is required solely for the considered object, not the environment, but it would be
a good improvement, especially for deformable objects during animations.

\bibliographystyle{abbrv}
\bibliography{fabrizio_col_det}

\end{document}